\documentclass[10pt,conference,letterpaper]{IEEEtran}

\IEEEoverridecommandlockouts
\usepackage{cite}
\usepackage{amsmath,amssymb,amsfonts}
\usepackage{algorithmic}
\usepackage{graphicx}
\usepackage{textcomp}
\usepackage{xcolor}
\usepackage{xspace}
\usepackage{booktabs}
\usepackage{multirow}
\usepackage{siunitx}
\usepackage{url}

\newcommand{\Gazebo}{Gazebo\xspace}
\newcommand{\Mujoco}{MuJoCo\xspace}
\newcommand{\Pybullet}{PyBullet\xspace}
\newcommand{\Webots}{Webots\xspace}

\newcommand{\DLR}{German Aerospace Center (DLR)\xspace}
\newcommand{\ZAL}{ZAL Center of Applied Aeronautical Research\xspace}

\def\BibTeX{{\rm B\kern-.05em{\sc i\kern-.025em b}\kern-.08em
    T\kern-.1667em\lower.7ex\hbox{E}\kern-.125emX}}

\usepackage{tikz}
\usepackage{pgfplots}
\pgfplotsset{compat=newest}
\usetikzlibrary{plotmarks}
\usetikzlibrary{arrows.meta}
\usepgfplotslibrary{patchplots}
\usepackage{grffile}

\usepackage{tikzscale}
\usetikzlibrary{calc}

\usetikzlibrary{positioning}
\usepgfplotslibrary{polar}
\usepgfplotslibrary{groupplots}
\usetikzlibrary{patterns} 
\pgfplotsset{plot coordinates/math parser=false}
\pgfplotsset{ylabsh/.style={every axis y label/.style={at={(0,0.5)}, xshift=#1, rotate=90}}}

\usepackage{geometry}
\renewcommand{\IEEEtitletopspaceextra}{\dimexpr-\headheight-\headsep+18pt\relax}

\newcommand\copyrighttext{%
  \footnotesize \textcopyright~2021 IEEE.
  Personal use of this material is permitted.
  Permission from IEEE must be obtained for all other uses, in any current or future media, including reprinting/republishing this material for advertising or promotional purposes, creating new collective works, for resale or redistribution to servers or lists, or reuse of any copyrighted component of this work in other works.}
\newcommand\copyrightnotice{%
\begin{tikzpicture}[remember picture,overlay]
\node[anchor=south,yshift=10pt] at (current page.south) {\fbox{\parbox{\dimexpr\textwidth-\fboxsep-\fboxrule\relax}{\copyrighttext}}};
\end{tikzpicture}%
}

\begin{document}

\title{\vspace{26pt}Comparing Popular Simulation Environments in the Scope of Robotics and Reinforcement Learning\\
\thanks{This work is partially funded through the German Federal R\&D aid scheme for the aeronautics sector - (LuFo VI) by the Federal Ministry of Economic Affairs (BMWi), supported by the Project Management Agency for Aeronautics Research (PT-LF), a division of the German Aerospace Center (DLR). Project 20D1926: Artificial Intelligence Enabled Highly Adaptive Robots for Aerospace Industry 4.0 (AIARA).}
}

\author{
\IEEEauthorblockN{%
Marian Körber\IEEEauthorrefmark{1},
Johann Lange\IEEEauthorrefmark{2}\textsuperscript{\textsection},
Stephan Rediske\IEEEauthorrefmark{2}\textsuperscript{\textsection},
Simon Steinmann\IEEEauthorrefmark{1}\textsuperscript{\textsection},
Roland Glück\IEEEauthorrefmark{1}}
\IEEEauthorblockA{\IEEEauthorrefmark{1}\textit{Institute of Structures and Design}, \IEEEauthorrefmark{2}\textit{Innovation Services} \\
\IEEEauthorrefmark{1}\textit{\DLR}, \IEEEauthorrefmark{2}\textit{\ZAL}\\
\IEEEauthorrefmark{1}Augsburg, Germany, \IEEEauthorrefmark{2}Hamburg, Germany \\
\{marian.koerber, roland.glueck\}@dlr.de, \{johann.lange, stephan.rediske\}@zal.aero, simon.steinmann91@gmail.com}
}

\maketitle
\copyrightnotice
\begingroup\renewcommand\thefootnote{\textsection}
\footnotetext{These authors contributed equally and are ordered alphabetically.}
\endgroup

\begin{abstract}
    This letter compares the performance of four different, popular simulation environments for robotics and reinforcement learning (RL) through a series of benchmarks.
    The benchmarked scenarios are designed carefully with current industrial applications in mind.
    Given the need to run simulations as fast as possible to reduce the real-world training time of the RL agents, the comparison includes not only different simulation environments but also different hardware configurations, ranging from an entry-level notebook up to a dual CPU high performance server.
    We show that the chosen simulation environments benefit the most from single core performance.
    Yet, using a multi core system, multiple simulations could be run in parallel to increase the performance.
\end{abstract}


\section{Introduction}
    In the last years, simulations became an ever more important part of hardware development, especially in the field of robotics and reinforcement learning (RL) \cite{Ivaldi2014, Kober2013}.
    Modern RL methods show incredible results with virtual humanoid models learning to walk, robots learning to grip and throw different objects, to name a few \cite{Tassa2018, Polydoros2017, Erickson2019}.
    Many of these use cases describe theoretical fields of application and are not yet used in a value adding environment.
    In fact, however, the RL methodology has great advantages in the field of industrial production automation.
    Especially the task of automated process control can benefit from the versatility and flexibility of an RL agent, mostly because current process flows in industrial environments are based on fixed programmed algorithms, offering limited flexibility to react to changed process states.
    In particular for complex processes, fixed programmed algorithms quickly reach the limits of their capabilities simply because the integration of experience into classic code is often only possible to a restricted extent.
    RL has the ability to learn from experiences and apply the learned knowledge to real production processes.
    Similar to other machine learning applications, the acquisition of training data is one of the more challenging tasks.
    The quality of training data closely correlates to the quality of the simulation environment generating this data. 
    Thus, the quantity of the training data depends on the speed of the simulation environment.
    
    Therefore, developers need a reliable simulation environment with a physical model which represents their process environment well enough to produce meaningful data \cite{Denil2016}.
    Throughout the years, a large number of physics-based simulation environments that meet those requirements has been developed. 
    Especially developers of automation applications or scientists who have no previous experience with simulation environments and physic engines find it difficult to choose the right environments for their projects.
    
    The goal of this work is to facilitate the selection of simulation environments by presenting evaluations of simulation stability, speed, and hardware utilization for four simulation environments which are widely used in the research community.
    The focus of the investigations is placed on the realization of RL applications. 
    This also means that, in comparison to the classical development of algorithms, minor inaccuracies in the calculation of physical effects can be tolerated. 
    Advantages as well as disadvantages will be highlighted which help to emphasize the applicability of the selected simulation environments in the industrial context. 
    More precisely, criteria such as usability, model creation and parameter setup have been evaluated.
    All the APIs used for this work are Python-based.
    
    In addition, a statement is made regarding which of the selected tools can be used to create a stable simulation most quickly and easily by an inexperienced user. 
    However, as there is no one-fits-all solution, we cannot give a single, universal recommendation about which simulation environment to use.
    Instead, this paper is intended to facilitate the selection of RL methods especially for scientists and developers with little previous experience in this field.
    
    From the beginning, the authors had basic experiences with diverse simulation environments which are described in section \ref{s:SimEnvs}.
    During the development of the benchmark, they intensively worked with each of the four simulation environments considered in this work.
    This has resulted in a profound knowledge of the simulation applications even though it cannot be equated with the knowledge of a long-standing expert. 
    We are aware that the results and estimates for individual simulation environments could be refined and optimized by such an expert. 
    However, the data and analyses of this work correspond better to the working reality of beginners or advanced users of the considered simulation environments.
    The individual simulation environments use existing or customized physic engines to calculate the behavior of bodies with mass under the influence of forces such as gravity, inertia or contact constraints.
    In this work, we focus on the applicability of the simulation tool but not on the accuracy of the physic engines since this represents a separate field of research.
    None of the authors has any business or financial ties to the developers or companies of the simulation environments discussed here.
    A detailed description of the scenarios, parameters and settings used can be found at \url{https://github.com/zal/simenvbenchmark}.

\section{Related work}
    With the increasing popularity of RL applications, several papers have focused on the capabilities and functions of physics engines. 
    The emphasis of our work is on different aspects.
    
    Ayala et al. \cite{Ayala2020} focused on a quantitative, hardware load specific comparison of \Webots, \Gazebo and V-Rep.
    They were using the humanoid robot NAO running around a chair controlled by an external controller. 
    The work evaluated the system load of each simulation environment on one low-end PC system without comparing it to the real-time factor ($RTF$).
    
    Pitonakova et al. \cite{Pitonakova2018} compared \Gazebo, V-Rep and ARGoS for their functionalities and simulation speed. 
    Their work focuses on the development of mobile robot controls.
    Therefore, two model scenes were used to perform the benchmarks in GUI and headless mode.
    The evaluation includes the $RTF$, CPU and memory utilization.
    They were able show that \Gazebo is faster for larger model scenes whereas ARGoS is able to simulate a higher number of robots in smaller scenes.
    Unfortunately, the used benchmark models are not identical from simulator to simulator. 
    Instead, mainly already existing models from the internal libraries are used which differ from each other, making the objective comparison more difficult.
    
    The developers of \Mujoco describe a slightly different approach compared to the work mentioned above \cite{Erez2015}.
    They compare abstract, dynamic application scenarios, based on physical engines (Bullet, Havok, \Mujoco, ODE, and PhysX) on which the simulation environments are based on.
    These engines are being evaluated in terms of simulation stability and speed by increasing the time step after each run.
    Thus, it is possible to determine at which time steps errors occurred compared to the runs at the lowest time step.
    Such a measure is, in general, a good approximation at which time step the simulation becomes unstable.
    These findings rate \Mujoco as the most accurate and fastest physic engine regarding robotic applications.
    The other engines show their potential in game scenarios and multi-body models with unconnected elements.

\section{Objectives of the Benchmark}
\label{s:Objectives}
\begin{figure}[htbp]
        \centering
        \includegraphics[trim=0 100 0 40, clip, width=\linewidth]{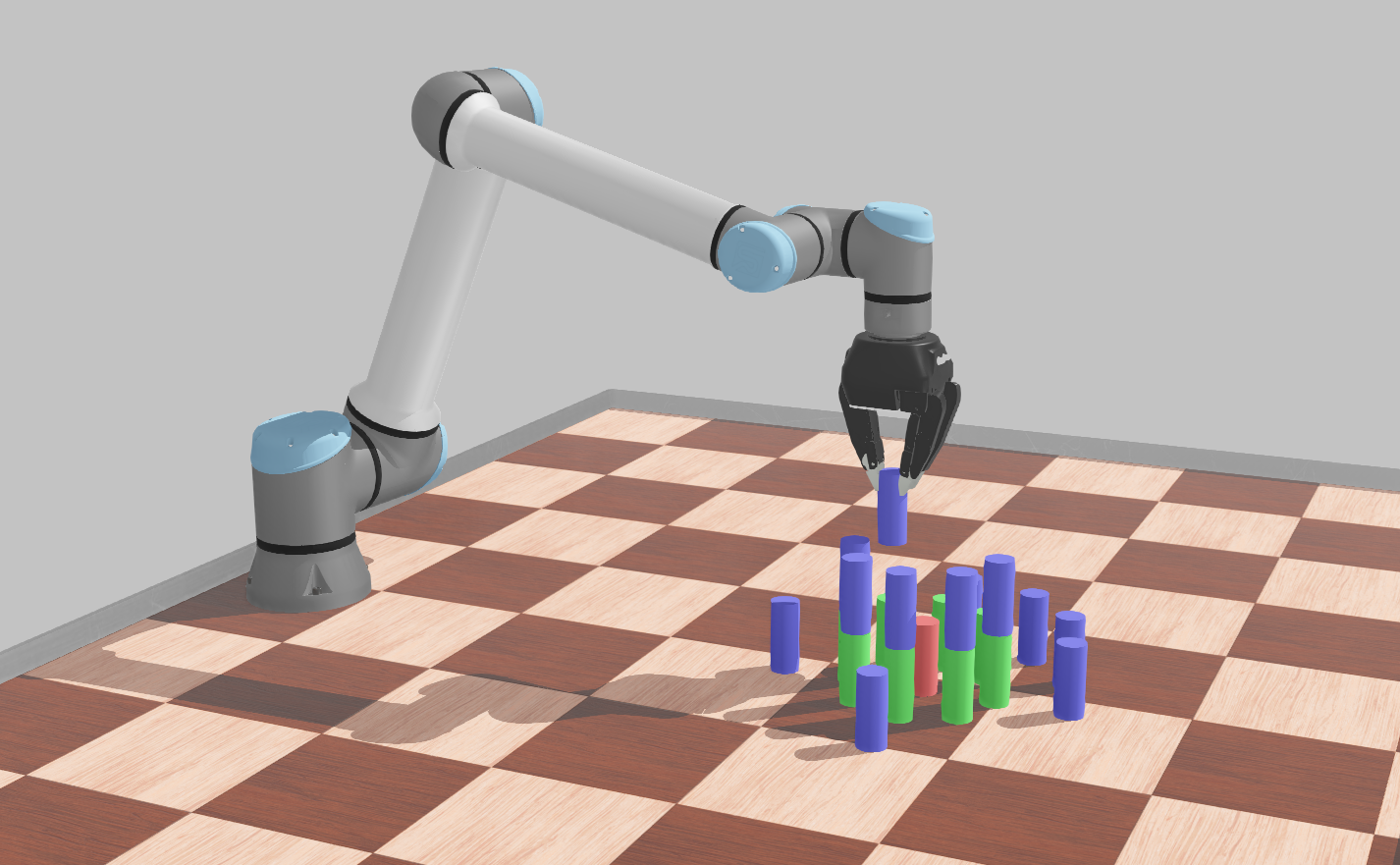}
        \caption{The first scenario with the robot and the cylinders in \Webots.}
        \label{fig:Scenario:robot}
    \end{figure}

\begin{figure}[htbp]
        \centering
        \includegraphics[trim=0 60 0 100, clip, width=\linewidth]{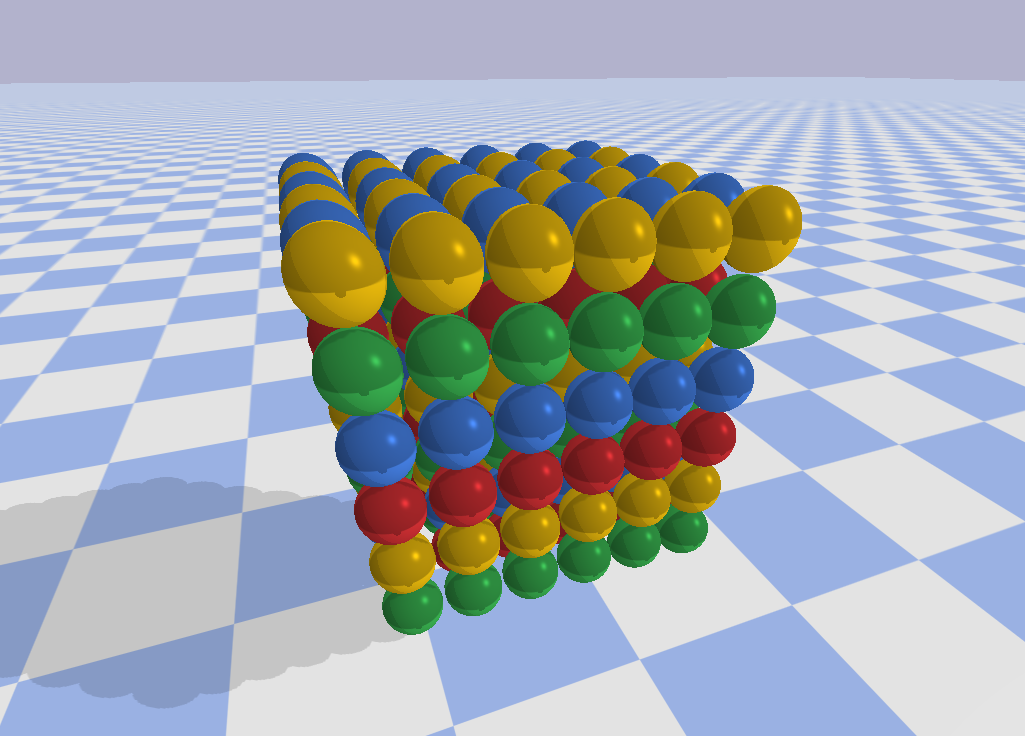}
        \caption{The second scenario with a cube build from spheres in \Pybullet.}
        \label{fig:Scenario:spheres}
    \end{figure}

    Given our need for a fast, yet stable and precise simulation environment, we decided to build two scenarios.
    The first scenario is closely related to our use case, simulating industrial robots.
    Thus, we build the scenario around the Universal Robot UR10e equipped with the Robotiq 3-Finger Gripper, a commonly used robot and endeffector, for example in \cite{Korkmaz2018}.
    In this scenario the robot was given the task to rearrange and stack small cylinders, as seen in Fig.~\ref{fig:Scenario:robot}.
    To achieve the task, the robot, especially the complex gripper, has to perform small and precise movements.
    Likewise, the simulation environment has to be precise for the task not to fail.
    However, this scenario is characterized by a low number of simultaneous contacts.
    As many scenarios, especially in industrial applications, consist of multi-body-collisions, such as a robot sorting through a bin of objects, we designed a second benchmark scenario.
    The second rather generic scenario is built around a multitude of spheres.
    Those are arranged in a cube with side length of \num{6} spheres, see Fig.~\ref{fig:Scenario:spheres}, a total of \num{216} densely packed spheres, all falling down at the same time and then spreading frictionless on the floor, generating a higher number of simultaneous contacts.

\section{Considered Simulation Environments}
\label{s:SimEnvs}
    Among the most popular rigid body simulation environments for robotics and RL are \Gazebo, \Mujoco, \Pybullet, and \Webots \cite{Ivaldi2014, Staranowicz2011}.
    However, while each physics engine is designed to portray the real world in a general way, each implementation has its own strengths and caveats \cite{Millington2007}.
    
    In contrast to simulations used in finite element method (FEM) and computational fluid dynamics (CFD), which can take hours or days for a single simulation step, robotic and RL applications require a responsive simulation environment running at least at real-time.
    Responsiveness can be achieved when the $RTF$, the ratio of simulation time over real-world time, is at least around \num{1}.
    Furthermore, especially RL greatly benefits from a $RTF$ of above \num{1}, as the agent can now be trained faster than in real-time leading to lower waiting times for the training.
    Hence, while pure robotic simulations like a digital-twin in general do not benefit from an $RTF$ above \num{1}, RL does.
    Nevertheless, higher real-time factors always come with lower precision as we will show in section \ref{s:Results}.
    Thus, as FEM and CFD need to be as precise as possible, they are accordingly slow.
    Robotic and RL application in most cases do not need such precision and can thus use simulation environments with reduced precision in favor of speed.
    A simple and often used way to increase the speed is by increasing the time step of the simulation, simultaneously reducing the precision.
    However, the precision needs to be good enough, meaning the simulation must portray the real world physics good enough for the individual use case to work.
    In the presented scenario regarding the UR10e with Robotiq 3-Finger Gripper, all simulators, apart from \Gazebo, used a joint position controller.
    Since the existing repositories for \Gazebo used an joint trajectory controller with effort interface, we chose to use it as well. 
    For this, the PID values had to be tuned, which introduced an additional source of error.

    \subsection{\Gazebo}
    \label{ss:Gazebo}
        \Gazebo is a open-source robotics simulation environment and supports four physics engines: Bullet \cite{pybullet}, Dynamic Animation and Robotics Toolkit (DART) \cite{Lee2018}, Open Dynamics Engine (ODE) \cite{ode} and Simbody \cite{simbody}.
        The development started in 2002 at the University of Southern California as a stand-alone simulator and was expanded in 2009 with an integration of an interface for the Robot Operating System (ROS).
        Since the foundation of the Open Source Robotics Foundation (OSRF) \cite{open_robotics} in 2012, OSRF has been leading the development and is supported by a large community \cite{gazebo}.
    
    \subsection{\Mujoco}
    \label{ss:MuJoCo}
        \Mujoco (Multi-Joint dynamics with Contact) is a simulation environment and physic engine focused on robotic and biomechanic simulation, as well as animation and machine learning application \cite{Todorov2012}. 
        It is commonly known for RL applications that train an agent to enable virtual animal or humanoid models to walk or perform other complex actions \cite{Booth2019a}.
        \Mujoco provides a native, XML-based model description format which is designed to be human readable and editable.
        In contrast to the other simulation environments described in this work, a license is required to be able to install it. However, not all licenses allow the usage of \Mujoco inside a Docker container.
        Only an academic license allows using \Mujoco inside a Docker container which we were not able to acquire.
    
    \subsection{\Pybullet}
    \label{ss:pybullet}
        The simulation environment \Pybullet is based on the Bullet physics-based simulation environment.
        It focuses on machine learning applications in combination with robotic applications \cite{Coumans}.
        \Pybullet is characterized in particular by a large community \cite{pybullet_community} which further develops this simulation environment as an open-source project and offers support for beginners. 
        In addition, the import of robot and machinery models is simplified as a wide variety of model formats can be loaded, such as SDF, URDF and MJCF (\Mujoco's model format). 
    
    \subsection{\Webots}
    \label{ss:Webots}
        \Webots is an open source and multi-platform desktop application used to simulate robots. It provides a complete development environment to model, program and simulate robots \cite{Webots}.
        \Webots offers an extensive library of sample worlds, sensors and robot models.
        Robots may be programmed in C, C++, Python, Java, MATLAB or ROS, with detailed API documentation for each option.
        Models can be imported from multiple CAD formats or converted from URDF.
        At its core, \Webots uses a customized ODE physics engine.

\section{Benchmark Scenarios}
\label{s:Benchmarks}
    As already mentioned in section \ref{s:Objectives}, we defined two scenarios for this benchmark.
    The first benchmark is built around the Universal Robot UR10e robot equipped with the Robotiq 3-Finger Gripper.
    It is commonly used in industrial applications, especially for grasping tasks \cite{Korkmaz2018, Sadun2017}.
    In this scenario, we model the grasping task by rearranging and stacking multiple cylinders, as shown in Fig.~\ref{fig:Scenario:robot}.
    In total, there are \num{21} cylinders, of which the robot directly interacts with \num{9} cylinders by gripping those.
    The first task of the robot is to stack one cylinder on top of another \num{8} times along a circle.
    Afterwards, the robot builds a tower of \num{7} cylinders, the maximum it can stack whilst facing the hand downwards.
    The duration of the trajectory is \SI{279}{\second}.
    This scenario requires precise control of the UR10e and the gripper. The cylinders positions were chosen so that slight deviations of the position will knock over other cylinders.
    Moreover, if the grippers were unable to apply sufficient force to the cylinders they would slip and the task would fail.
    All those precise movements require the simulation environments to accurately calculate not only the state of the system but also the contacts between the objects.
    
    The second scenario is characterized by the increased number of simultaneous contacts that occur especially at the beginning of the benchmark.
    Here, multiple spheres are ordered in a cubical grid with side length of six spheres, see Fig.~\ref{fig:Scenario:spheres}.
    Once falling down, all those \num{216} spheres will interact with each other and start spreading frictionless on the ground floor.
    Due to the slight perturbations, which are the same for all simulations, in the initial position of the spheres along the grid the spheres spread randomly as opposed to a symmetrical spreading along the floor plane.
    After \SI{4}{\second}, the spheres are mostly moving freely on the ground plate and the simulation is stopped.
    As the aforementioned perturbations introduce chaos, we cannot clearly define the desired position for the spheres at any time past the moment, the spheres come into contact with one another.
    However, we can deduct if the spreading pattern itself is deterministic given a simulation environment and a time step.
    Thus, allowing us to qualitatively analyze the movement of the spheres and moreover determine the $RTF$ with respect to the number of theoretically contact, see section \ref{s:Results}.

\section{Software Framework}
    Each simulation environment has a specific controller, providing methods to change physics parameters, communicate with robot controllers and retrieve simulation data.
    The core of our software framework is a Python script, referred to as \texttt{taskEnv}, identical for each simulation environment with three main functions.
    First, change the time step of the connected simulation.
    Secondly, send identical control commands given the predefined trajectory. 
    And thirdly, retrieve data from the simulation and generate a log.
    A shell script launches the hardware monitor and each simulation environment executes the \texttt{taskEnv} for the specified time steps. 
    For the spheres scenario, the structure is virtually the same, with the difference of no trajectory control commands being transmitted.
    
    \subsection{Real-Time Factor Calculation}
    The real-time factor $RTF$ was calculated using $RTF = \Delta t_\mathrm{sim}\cdot\Delta t_\mathrm{real}^{-1}$, were $\Delta t_\mathrm{sim}$ is the duration of the current time step and $\Delta t_\mathrm{real}$ the time it took the system to calculate the respective time step.
    The total $RTF$ of one simulation run was then calculated as the average of all its $RTF$s.
    In \Mujoco, \Pybullet and \Webots, rather than continuously executing the simulation, each simulation step has to be explicitly called through the API.
    For these simulation environments $\Delta t_\mathrm{real}$ was calculated by taking a time stamp before and after each simulation step call using Python's \texttt{time.perf\_counter()}.
    For the average $RTF$ of a benchmark run the final $t_\mathrm{sim}$ was divided by the sum of all $\Delta t_\mathrm{real}$. 
    
    Since we wanted to rely on the existing Phyton APIs and did not want to write any further plugins, we had to use the existing control mechanisms, such as the ROS API for \Gazebo, namely rospy.
    While \Gazebo also allows to call each step explicitly, rospy does not.
    Therefore, \Gazebo runs continuously and the $RTF$ was calculated by taking a timestamp at the start and end of a benchmark run.
    Simulation time $t_\mathrm{sim}$ was retrieved by subscribing to the ROS topic \texttt{/clock}.
    This may lead to inaccurate results for the significantly shorter sphere scenario which only lasts for four simulation seconds, especially for higher time steps.
    Furthermore, it should be noted that our $RTF$ values do not necessarily represent the $RTF$ values displayed by the GUIs of each simulation environment as different formulas might be used. 
    
\section{Benchmark Procedure}
    As discussed before, the speed of the simulation environment is heavily dependent on the time step.
    The same applies to precision, but anti-proportional.
    Thus, we ran the benchmark for different time steps, evaluating the performance and, in the case of the robot scenario, also the precision.
    In our setup, the time steps range from \SIrange{1}{64}{\milli\second}, starting in steps of \SI{1}{\milli\second}.
    Moreover, in order to gain statistical significance, we ran each time step \num{5} times per scenario, environment and hardware configuration.
    Furthermore, we ensured the CPU cooled down again after every run to not run into thermal throttling which could reduce the performance and influence the results for subsequent simulation environments.
    This way, we wanted to ensure that we did not bias the results given the order we ran the benchmarks.
        
    \subsection{Hardware Systems Used for Benchmark Run}
    To reflect the wide variety of possible systems used to compute the simulations, multiple target systems were defined.
    The configuration was chosen to reflect typical systems available in research for simulating and training an environment.
    Thus, the configurations range from an entry-level notebook with only an Intel iGPU to a rendering/simulation server, which might be available at an institution or can be rented at Amazon AWS or Microsoft Azure.
    The technical specifications are listed in table \ref{tb:hardware_systems}.
    \begin{table}[htbp]
    \caption{Hardware Systems}
    \begin{center}
    \begin{tabular}{
    @{}
    >{\raggedright\arraybackslash}p{0.15\linewidth}
    >{\raggedright\arraybackslash}p{0.16\linewidth}
    >{\raggedright\arraybackslash}p{0.24\linewidth}
    >{\raggedright\arraybackslash}p{0.09\linewidth}
    >{\raggedright\arraybackslash}p{0.13\linewidth}
    @{}}
    \toprule
     & CPU & GPU & RAM & Storage \\\midrule
    Server & $\num{2}\times$ AMD EPYC 7542 & $\num{4}\times$ NVIDIA Quadro RTX 8000 & \SI{512}{\gibi\byte} DDR4 & Samsung PM1733 \\
    Mobile workstation & Intel Core i7-8700 & NVIDIA GeForce RTX 2080 Mobile & \SI{32}{\gibi\byte} DDR4 & Samsung 960 EVO \\
    Notebook & Intel Core i7-7500U & Intel HD Graphics 620 & \SI{8}{\gibi\byte} DDR4 & Samsung 950 Pro \\
    \bottomrule
    \end{tabular}
    \label{tb:hardware_systems}
    \end{center}
    \end{table}
    All systems were up-to-date as of the date of the benchmarks with the host OS being Ubuntu 20.04.1 LTS with kernel 5.4.0-52-generic, Docker version 19.03.13 (build 4484c46d9d), Python 3.8.2, and ROS Noetic Ninjemys.
    The systems with an NVIDIA GPU both used driver version 455.23.04 coupled with CUDA 11.1, as well as NVIDIA-Docker version 2.5.0-1 to enable GPU pass-through.
    The simulation environments were running versions 11.1.0 for \Gazebo, 200 for \Mujoco, 2.8.4 for \Pybullet, and R2020b-rev1 for \Webots.
    
    While a GUI is important for developing the scenario in every simulator, we chose to run the simulations "headless", i.e. no GUI was spawned.
    This decision was based on the fact that many servers applications are running headless.
    Running the simulation environments headless is a built-in option for \Gazebo, \Mujoco and \Pybullet.
    Only \Webots does not currently offer support for running headless. 
    However, it provides a mode called "fast", which does not render any visualisation while still spawning a GUI.
    Therefore, we used Xvfb (X Window Virtual Framebuffer) which creates a virtual X-server entirely on the CPU for the \Webots GUI to attach to.
    But as the GUI is not rendering anything, this does not affect the performance.
    
    \subsection{Hardware Load Acquisition}
    \label{hardware_load_acqui}
        One of the main aspects of a high performance simulation is to what extent the simulation environment itself is able to utilize the hardware resources. 
        Therefore, the hardware load is monitored during every simulation run.
        To prevent the monitoring process from affecting the simulation performance, the software collectl \cite{collectl} is used in a separate thread.
        It is able to collect a wide variety of system data.
        In this work we will put focus on the CPU load to give an idea how big the hardware requirements of each simulation environments are.

    \subsection{Container-based Software Deployment}
    \label{ss:Docker}
        As the goal of this paper is to run the benchmark on as many different systems as possible, we chose Docker due to its easy portability.
        The usage of Docker enables two important use cases for this benchmark.
        Firstly, it lets us create an easy to deploy image with all necessary software already installed which also runs the full benchmark setup upon starting the container.
        Secondly, due to Docker's nature of userspace isolation \cite{Merkel2014}, we create a clean environment for our simulations as well as the hardware load acquisition to run.
        According to multiple studies \cite{Patel2018, Joy2015, Kovacs2017, Chae2019}, the performance loss of Docker is negligible, especially for CPU, GPU and RAM access.
        Despite the clear advantage of using Docker, we were not able to use \Mujoco inside the Docker container due to licensing problems, see section \ref{ss:MuJoCo}.

\section{Simulation and Results}
\label{s:Results}
    \subsection{Simulation Stability and Performance}
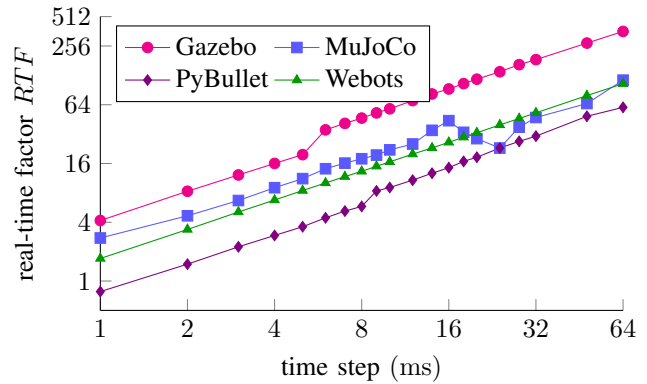
\begin{figure}[htbp]
        \centering
        \begin{tikzpicture}

\begin{axis}[%
width=3.566in,
height=3.566in,
at={(1.236in,0.481in)},
scale only axis,
xmin=1,
xmax=64,
xtick={1,2,4,8,16,32,64},
xlabel style={font=\color{black}},
xlabel={time step $(\si{\milli\second})$},
ymin=0.5,
ymax=512,
ytick={1, 4, 16, 64, 256, 512},
ylabel style={font=\color{black}},
ylabel={real-time factor $RTF$},
axis background/.style={fill=white},
axis x line*=bottom,
axis y line*=left,
width=0.8\linewidth,
height=0.45\linewidth,
legend pos=north west,
legend columns=2,
legend cell align=left,
xmode=log,
ymode=log,
log basis x={2},
log basis y={2},
scaled ticks=false,
xticklabel style=log ticks with fixed point,
yticklabel style=log ticks with fixed point
]

\addplot [color=magenta, mark=*] table [x index=0, y index=1]{
1.0 4.169580684841292 3.9809211609256243 4.282676695918166
2.0 8.311505015292866 8.072257891342094 8.628353767787797
3.0 12.208906007730038 11.598551766394442 13.02077591245927
4.0 16.02581435962921 15.242400253704343 16.72791314407999
5.0 19.722012729622833 19.571007808445522 19.89865681566993
6.0 35.39268813006106 34.26115237680097 35.88262534387071
7.0 41.2253201050046 40.70371626442403 41.97837405506484
8.0 46.71590408346479 46.431232543888235 47.398094713835455
9.0 52.81160321073529 51.95276961010252 53.680783997733315
10.0 58.12973159745117 57.60909021020993 58.60866048895759
12.0 70.55232101620274 68.80571288812477 71.63094621837101
14.0 82.4928998075514 81.52245568083994 83.59038127651381
16.0 92.84440528156672 91.30260216314434 94.59225251085333
18.0 105.29397897537869 102.85746269060023 107.27240259067139
20.0 117.0097426061377 115.64665202995772 118.24577594243642
24.0 139.37051976344304 136.19598901624505 142.22558682604205
28.0 164.88443008883328 159.49521735440885 170.09252337103698
32.0 185.56421179027961 180.90341947176714 192.85275525351614
48.0 274.22988819171803 264.20916499711853 284.1190069539228
64.0 360.87733867446076 346.56485324136185 378.66079926215156
};
\addlegendentry{\Gazebo}

\addplot [color=blue!65!white, mark=square*] table [x index=0, y index=1]{
1.0 2.759777471415656 2.726609721629533 2.78286872623387
2.0 4.6612548984743025 4.6550651582192595 4.6656138542179075
3.0 6.682989328290813 6.67692641161956 6.686776173119882
4.0 9.037076317131481 9.015262143431169 9.051684222950707
5.0 11.198909632811759 11.10556624155462 11.24325784981588
6.0 14.142794861979306 14.061851955117834 14.194999080541638
7.0 16.13923512374638 16.109220131614276 16.184404200441424
8.0 17.895139139645735 17.83939657813052 17.947390946055677
9.0 19.514295135779825 19.400934882897687 19.69049536886671
10.0 22.07042053491807 21.785956863262598 22.208331559984906
12.0 25.3357589728773 25.281645355270275 25.446860096076733
14.0 35.00216995108006 34.96403768431131 35.10504896140691
16.0 43.96043890498466 43.85398961589194 44.04200174446752
18.0 33.40016448772033 33.35962722098991 33.43074713763181
20.0 28.77138323459014 28.69337604059938 28.839436278008595
24.0 23.037808100064943 22.930525375789337 23.118511876123524
28.0 37.73072895975692 37.64905716436412 38.01664564559972
32.0 47.344221872419446 47.000591999915066 47.54305826079398
48.0 65.8744909323497 65.65319651169474 66.32952270198254
64.0 114.243859053078 113.88925821604218 114.65262130101053
};
\addlegendentry{\Mujoco}

\addplot [color=violet, mark=diamond*] table [x index=0, y index=1]{
1.0 0.7794056768346392 0.776431179152615 0.7835654754836303
2.0 1.489041843458437 1.4821425821068237 1.4963882856818669
3.0 2.240697567878528 2.2327825962465364 2.2469976616660245
4.0 2.929875913584188 2.9272110170917505 2.933673100706632
5.0 3.600820366923982 3.590649714448261 3.6108574758770176
6.0 4.442442555409107 4.387863888306789 4.46656942853094
7.0 5.207251774519625 5.189762594181027 5.218261912354556
8.0 5.812052626355139 5.783391086393293 5.830350862959668
9.0 8.377846492846093 8.366401546108653 8.38431470066138
10.0 9.082550588212722 9.058268052270218 9.109296947220125
12.0 10.804366250457454 10.768749613090893 10.842250021584075
14.0 12.72108725438749 12.67882073412837 12.765737288211799
16.0 14.498241018816554 14.318905499890086 14.71383846492655
18.0 16.816129126960327 16.762340194378886 16.881022699260058
20.0 18.553005671710444 18.44080769264593 18.65821837217409
24.0 23.06593446590893 22.977640345305378 23.139263364460238
28.0 26.97176944341955 26.93987450422128 27.062371388089385
32.0 30.51544377868937 30.471416522288724 30.617136972162157
48.0 48.74973047631703 48.66806999148147 48.79914121642411
64.0 60.20517697192618 59.859256018024 60.350324923173986
};
\addlegendentry{\Pybullet}

\addplot [color=green!60!black, mark=triangle*] table [x index=0, y index=1]{
1.0 1.7023163575919409 1.6903406820471996 1.707954050378208
2.0 3.3752374971169345 3.363499156575564 3.3803983794131707
3.0 5.095199615258373 5.0883363939411135 5.102111553263376
4.0 6.759821032581539 6.702481974552964 6.794790872239637
5.0 8.451057737586916 8.4160390744617 8.483853958160909
6.0 10.113911780611687 10.084369575538235 10.150054311453067
7.0 11.740822577593512 11.709641554365428 11.782950756090818
8.0 13.306804950211387 13.199943340756157 13.442232906223884
9.0 15.00075288121569 14.920387114336284 15.09309318777129
10.0 16.639510904043096 16.54342069078517 16.830841059077798
12.0 19.972112198704238 19.883503921682046 20.01561437614542
14.0 23.10654659980548 22.92500544533157 23.24275196568541
16.0 26.386872512658208 26.286680579342438 26.54950087902146
18.0 29.757993182359275 29.62166151191924 29.941505986595594
20.0 33.05745599235147 32.86768552534037 33.22570903257083
24.0 39.770858256138624 39.409715486180446 40.143263909819616
28.0 46.24357906080174 45.97837100574074 46.39770286924349
32.0 52.88637194912739 52.08386058280232 53.41742749775579
48.0 79.46456089707169 78.9701493886841 79.82327146499735
64.0 105.18030017776307 104.46138656606266 105.58587017276099
};
\addlegendentry{\Webots}

\end{axis}
\end{tikzpicture}%
        \caption{Progression of the $RTF$ for the robot scenario over increasing time steps running on the workstation.}
        \label{fig:rtf_RobotSim}
    \end{figure}

\begin{figure}[htbp]
        \centering
        \begin{tikzpicture}

\begin{axis}[%
width=3.566in,
height=3.566in,
at={(1.236in,0.481in)},
scale only axis,
xmin=1,
xmax=64,
xtick={1,2,4,8,16,32,64},
xlabel style={font=\color{black}},
xlabel={time step $(\si{\milli\second})$},
ymin=0.5,
ymax=32,
ytick={0.5,1,2,4,8,16,32},
ylabel style={font=\color{black}},
ylabel={real-time factor $RTF$},
axis background/.style={fill=white},
axis x line*=bottom,
axis y line*=left,
width=0.8\linewidth,
height=0.45\linewidth,
legend pos=south east,
legend columns=2,
legend cell align=left,
xmode=log,
ymode=log,
log basis x={2},
log basis y={2},
scaled ticks=false,
xticklabel style=log ticks with fixed point,
yticklabel style=log ticks with fixed point
]

\addplot [color=magenta, mark=*] table [x index=0, y index=1]{
1.0 1.3177370701931341 1.1463911553624435 1.3978473788618857
2.0 2.642455613527385 2.3203475740524855 2.772292829140825
3.0 3.3626003908793862 2.8973492754687205 3.832555873455322
4.0 4.253012358116002 3.6352599672054344 4.632957271237935
5.0 4.996692161695291 3.5340253110218867 5.841220267276926
6.0 6.765471609439291 5.4744117650926665 7.896106089859658
7.0 6.479665716934572 4.1517228920880225 8.15256745959101
8.0 7.523639091563436 5.705649646031029 9.168918105670176
9.0 7.38362217154275 5.925856462417925 9.994236164837744
10.0 7.073305523806877 3.962841409083671 10.736414525770828
12.0 8.692245125976058 3.165505384603361 15.091843186477552
14.0 10.22287880119337 3.2225400441613195 18.976305590951394
16.0 9.71557323420508 4.908860891954783 18.88526225949746
18.0 11.630162781650125 3.490517106317696 20.910530951391774
20.0 8.470382526017087 1.73057813343325 12.962931352971879
24.0 12.35760507696573 4.303261519130681 19.34122478349659
28.0 9.871025324940284 1.5178006086026812 16.06590239914683
32.0 25.496564479347512 16.02121015313934 41.628402834798834
48.0 26.251786545711695 3.5685105971770272 43.41074944916788
};
\addlegendentry{\Gazebo}

\addplot [color=blue!65!white, mark=square*] table [x index=0, y index=1]{
1.0 0.7807067932196305 0.7799682915887486 0.7827904928329699
2.0 1.587753434198571 1.5853875204304397 1.5892041007303568
3.0 2.2169543326969317 2.2051464498683577 2.225857030492742
4.0 2.9760172847347293 2.9739315781680458 2.979882792429254
5.0 3.6906794245799133 3.6817402473924634 3.698319119740068
6.0 4.407807932929358 4.373803108819584 4.423045780467146
7.0 5.120676968172125 5.097990123635672 5.1376059973221855
8.0 6.107068989256651 6.091768672113307 6.117654247229233
9.0 7.245461098823073 7.2263764345741395 7.259979131630634
10.0 7.775249165894332 7.634835898996218 7.856290116157542
12.0 8.77643266418652 8.669742702506268 9.043351235499971
14.0 10.344449503991212 10.328927412559631 10.370530399549493
16.0 11.425989524521892 11.40209592880081 11.444321161458486
18.0 12.090345767794732 11.998603481911918 12.137618166640957
20.0 13.361191047355074 13.333470934955573 13.394068049541536
24.0 15.049385256476134 14.973127403168474 15.126114357453865
28.0 16.349684428356603 15.885305573422487 16.515433616968235
32.0 18.10039193618355 17.618136671636865 18.471056465331493
48.0 22.356180249021214 22.04193515359034 22.80421955361293
64.0 23.48816777604326 23.11678196727225 24.020564798195785
};
\addlegendentry{\Mujoco}

\addplot [color=violet, mark=diamond*] table [x index=0, y index=1]{
1.0 1.2802076921485592 1.2706177270221948 1.2855797270859046
2.0 2.2956648560920776 2.2913632357526628 2.301773821094973
3.0 3.3172109634668288 3.309269476727574 3.328141496456782
4.0 4.326004603796719 4.290011909828502 4.361744804382623
5.0 5.212757953635839 5.178838687991211 5.226081453344917
6.0 6.0701153437132715 6.039640276486223 6.105377264944536
7.0 7.037049424051259 7.004908519167922 7.069229044711056
8.0 8.009957154503898 7.9256236870168895 8.066784427622325
9.0 8.603843665752372 8.526470841840597 8.698833249178229
10.0 9.49629362907127 9.45691892956856 9.548154990731762
12.0 11.355162801306284 11.316937897990968 11.386920620535237
14.0 12.482615877072892 12.405167525594983 12.54821808004787
16.0 13.658133848280823 13.629147528423072 13.672153511728382
18.0 15.238793509866492 15.127982948248754 15.308932226988244
20.0 16.191629891999476 16.12679958449536 16.281552978195737
24.0 14.128803647692298 14.098488633427872 14.157911069677558
28.0 15.495849989772188 15.424585892379014 15.546466327342756
32.0 13.692211433518645 13.561331313987269 13.735625020542493
48.0 17.463289551390723 17.232048184542997 17.63017758632798
64.0 18.07551859781438 17.979842627294776 18.164248573355476
};
\addlegendentry{\Pybullet}

\addplot [color=green!60!black, mark=triangle*] table [x index=0, y index=1]{
1.0 1.179745963719185 1.1767489349325457 1.1813825135874312
2.0 2.213610582498045 2.189785341748285 2.2369938968625167
3.0 3.1498980703038106 3.1116972612337923 3.2072779643808613
4.0 3.9934515797688626 3.9443142974989005 4.0332729838540855
5.0 4.67605513557808 4.5913323487762465 4.734232900161574
6.0 5.2528651347461 5.211070354294715 5.2939008438147654
7.0 5.7088152929625355 5.623987103234838 5.807194994592705
8.0 6.152884277323032 6.12619752376017 6.167180007801441
9.0 6.449705331242953 6.419409479621715 6.4860624423007085
10.0 6.5277049721523595 6.505426552893817 6.576414670479367
12.0 6.824101765560513 6.759765063403776 6.948967879571286
14.0 6.9711930703418306 6.896530480665586 7.091370946152828
16.0 7.052770417401243 6.9994675851227495 7.159899269018742
18.0 6.782766142415954 6.574026483595128 6.868092199985555
20.0 7.036756479671425 6.952881167706969 7.119683661175165
24.0 6.70174558952773 6.647619076450003 6.765660675110574
28.0 6.017500033210672 5.942125262326591 6.086795532413194
32.0 4.63922013079866 4.611211576207055 4.684112631375234
48.0 3.745346345676243 3.6809798638274054 3.7724047865981114
64.0 3.4509111155706504 3.4253263197229047 3.470265954837816
};
\addlegendentry{\Webots}

\end{axis}
\end{tikzpicture}%
        \caption{Progression of the $RTF$ for the sphere scenario over increasing time steps running on the workstation.}
        \label{fig:rtf_nnnSim}
    \end{figure}
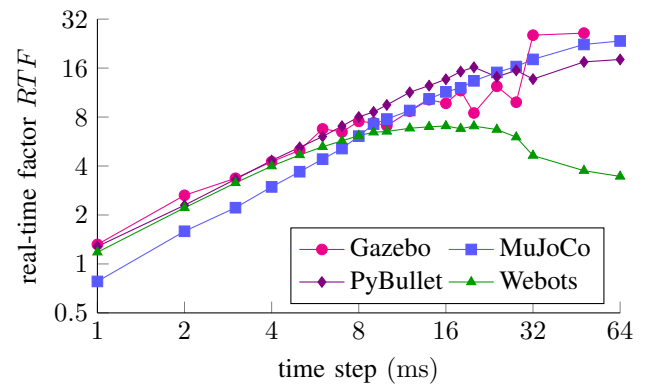

    Based on the data evaluation, the following describes which $RTF$ is achieved during the calculation of the two use cases. 
    Fig.~\ref{fig:rtf_RobotSim} shows the logarithmic progression of the $RTF$s over the also logarithmic increase of time steps for the robot scenario.
    The data indicates that the $RTF$s have a linear gradient over the time step. 
    However, it should be noted that the simulation does not provide valid results over the entire time step progression. 
    Due to the increasing time steps, errors occur during the physic calculations which are reflected in an incorrect reaction of the robot or the objects.
    We consider a simulation result as invalid if the deviation of a observed cylinder position compared to its target position is larger than \SI{1}{\centi\meter} as this would indicate it has slipped or fallen.
    Based on the solver parameters described in this thesis \Webots achieves a valid simulation result up to a time step of \SI{48}{\milli\second}.
    \Mujoco achieves valid results up to 3ms and \Pybullet up to \SI{7}{\milli\second}. 
    In the latter simulation environment, however, an outlier can be observed, since at 4ms a non-valid result is obtained.
    Apparently, the stacked tower collapses shortly before the end of the simulation.
    In \Gazebo all time steps are calculated, but the control of the Robotiq 3-Finger Gripper did not succeed to stabilize at time steps starting at \SI{2}{\milli\second}. 
    Thus, \Gazebo can only generate a valid result with time steps of \SI{1}{\milli\second} for the use cases discuss in this work.
    Fig.~\ref{fig:rtf_nnnSim} describes the same progression as Fig.~\ref{fig:rtf_RobotSim} but for the sphere scenario.
    Please note that \Gazebo did not generate any observations for \SI{64}{\milli\second}.
    For lower time steps, the same linear trend can be observed, but with a lower initial $RTF$ due to the increased number of contacts.
    Nevertheless, for higher time steps the stability of the simulation environments is not guaranteed leading to a break of the trend.
    In an in-depth analysis of the $RTF$ over the simulation time, we discovered that the $RTF$ is greater than the average $RTF$ at the beginning.
    As soon as the the spheres come into contact, the $RTF$ drops significantly, only to recover and flatten out once the spheres are distributed along the ground floor and interactions are rare.
    \subsection{Hardware Load}
    \label{s:hardware_load}
        The physics engines of the simulation environments are all designed to run only on the CPU.
        Except for \Gazebo and \Webots, no multi-threaded physics calculations are provided \cite{OpenSourceRoboticsFoundation2015, webots_multi}.
        Thus, \Mujoco and \Pybullet always only utilized a single core while \Webots and especially \Gazebo both can span specific tasks of their workload over multiple threads.
        Due to the comparability and the non-replicable simulations results \cite{Webots}, multi-threading functionalities are deactivated. 
        Therefore, we expected all simulation environment to only utilize one core up to \SI{100}{\percent}, especially because the simulation environments are set up to run headless with maximum $RTF$ possible. 
        \begin{table}[htbp]
        \caption{Average CPU Usage per Simulation Environment Compared to the Average Real-Time Factor $RTF$ with a time step of \SI{1}{\milli\second}.}
        \setlength{\tabcolsep}{0.8pt}
        \begin{center}
        \begin{tabular}{
        >{\raggedright\arraybackslash}p{0.13\linewidth}
        >{\raggedright\arraybackslash}p{0.10\linewidth}
        >{\raggedleft\arraybackslash}p{0.13\linewidth}
        >{\raggedleft\arraybackslash}p{0.10\linewidth}
        >{\raggedleft\arraybackslash}p{0.13\linewidth}
        >{\raggedleft\arraybackslash}p{0.10\linewidth}
        >{\raggedleft\arraybackslash}p{0.13\linewidth}
        >{\raggedleft\arraybackslash}p{0.10\linewidth}
        }
        \toprule
         & & \multicolumn{2}{c}{Server} & \multicolumn{2}{c}{Mobile workstation} & \multicolumn{2}{c}{Notebook} \\
         & & \multicolumn{1}{l}{CPU} & \multicolumn{1}{l}{$RTF$} & \multicolumn{1}{l}{CPU} & \multicolumn{1}{l}{$RTF$} & \multicolumn{1}{l}{CPU} & \multicolumn{1}{l}{$RTF$} \\\midrule
        \multirow[t]{2}{*}{\Gazebo} & robot & \SI{273.0}{\percent} & \num{4.3} & \SI{264.4}{\percent} & \num{4.4} & $(\SI{11.9}{\percent})$ & \num{2.3}  \\
                                    & spheres & \SI{223.1}{\percent}  & \num{1.1} & \SI{208.7}{\percent} & \num{1.4} & \SI{213.2}{\percent} & \num{0.8}  \\
        \multirow[t]{2}{*}{\Mujoco} & robot & \SI{116.3}{\percent} & \num{2.4} & \SI{100.8}{\percent} & \num{2.8} & \SI{103.1}{\percent} & \num{2.1} \\
                                    & spheres & \SI{113.4}{\percent} & \num{0.6} & \SI{99.8}{\percent} & \num{0.8} & \SI{102.8}{\percent} & \num{0.6} \\
        \multirow[t]{2}{*}{\Pybullet} & robot & \SI{119.3}{\percent} & \num{0.7} & \SI{102.0}{\percent} & \num{0.8} & \SI{104.0}{\percent} & \num{0.6} \\
                                      & spheres & \SI{118.0}{\percent} & \num{1.1} & \SI{100.7}{\percent} & \num{1.3} & \SI{103.8}{\percent} & \num{1.0} \\
        \multirow[t]{2}{*}{\Webots} & robot & \SI{124.3}{\percent} & \num{1.8} & \SI{105.6}{\percent}  & \num{1.7} & \SI{109.1}{\percent} & \num{1.3} \\
                                    & spheres & \SI{120.7}{\percent} & \num{1.2} & \SI{103.8}{\percent} & \num{1.2} & \SI{103.3}{\percent} & \num{0.5} \\
        \bottomrule
        \end{tabular}
        \label{tb:CPUusageRTF}
        \end{center}
        \end{table}
        The CPU load shown in Tab.~\ref{tb:CPUusageRTF} is calculated based on the cumulative CPU usage across all existing threads and on the average of the \SI{1}{\milli\second} runs of each simulation environment on one of the three hardware systems.
        During the robot scenario, the physics engine fully utilizes the capacity of one thread completely over the simulation run in all simulation environments except \Gazebo. 
        \Gazebo seems to distribute the calculation over several threads by default.
        Depending on the simulator, there is also some overhead for commands and observation acquisitions which are utilizing one or two additional threads to a small extent.
        The CPU load acquisition failed during \Gazebo \SI{1}{\milli\second} runs of the robot scenario on the notebook. 

\section{Conclusion}
    %
    The choice of the most suitable simulation environment depends strongly on the field of application of the respective RL use case.
    Therefore, the use cases described in section \ref{s:Objectives} were designed to reflect the broadest possible range of industrial applications. 
    Each of the four selected simulation environments were able to fulfill the tasks at low time steps.
    However, in order to accelerate the simulation speed and thus the potential learning process, the time step was increased.
    The chosen method of acceleration had a different impact on the quality of the results for each simulation environment.
    The data from section \ref{s:Results} shows that the stability and thus the reliability of the simulation environment is reduced by increasing the time step. The impact is considerable for most simulation environments but always for a different reason.
    While in \Gazebo the nine finger joints of the Robotiq 3-Finger Gripper became unstable, \Pybullet and \Mujoco became unstable due to contact calculations.
    
    \Webots, however, showed a surprisingly good stability in our investigations, up to a time step of \SI{48}{\milli\second}, retaining a high $RTF$. 
    Furthermore, \Webots offers a comprehensive GUI for creating and modifying the simulation model as well as extensive documentation. 
    Thus, \Webots is particularly suitable for developers in the field of industrial automation who are dealing with physic simulations for the first time and do not have the opportunity to familiarize themselves with solver optimizations.
    However, it should be noted that \Webots uses a specially customized ODE version and offers fewer configuration options compared to the other simulation environments. 

    \Mujoco is a highly optimized simulation environment that provides a wide range of solver parameters and settings. 
    Hence, this simulation environment can be adapted and optimized to any possible model. 
    With the settings we chose, it achieved a high $RTF$ in both simulation scenarios, whereby the simulation stability drops already at low time steps.
    \Mujoco is therefore aimed at experienced RL and robot developers who have the capability to work intensively with \Mujoco's own model language and the comprehensive setting options.
    
    \Pybullet is a sophisticated simulation environment that offers a wide range of functionality for creating, optimizing and controlling simulation models. 
    This includes model import functions for different formats such as SDA, URDF and MJCF, which simplifies the model generation.
    \Pybullet is based on an intuitive API syntax. In combination with extensive documentation \cite{pybullet} and a big community \cite{pybullet_community}, it allows developers to easily perform valid simulations, from the authors' point of view. However, a lower $RTF$ per time step compared to the other simulation environments needs to be accepted.
    This may partially be due to the default usage of the more accurate but slower cone friction model compared to the pyramid friction model used by the other simulation environments.
    However, the simulation results remain stable up to a time step of \SI{7}{\milli\second}.
    Therefore, \Pybullet is a simulation environment that is easy to learn and easy to use.
    
    \Gazebo, in combination with ROS, is a well-established simulation environment for the control and virtualization of robot applications.
    Unfortunately, we were not able to optimize the control of the Robotiq 3-Finger Gripper in a way that it behaves stable with a time step over \SI{1}{\milli\second}.
    The control concept of \Gazebo and ROS differs significantly from the concepts of other simulation environments that wait for a command from the controller before a simulation step.
    In \Gazebo, the simulation runs in parallel with ROS which provides the controllers for the robot. 
    In the provided packages, effort controllers were used which require meticulously tuned PID values.
    The PID values have been set to the best of our knowledge but are a potential cause of error.
    Although it is possible to use Gazebo for the training of RL agents, we cannot make any statement about the extent to which the simulation can be carried out over the time step of \SI{1}{\milli\second}.
    However, \Gazebo offers a large number of adjustable parameters.
    For example, the ODE solver can be switched from Quickstep (default) to Worldstep (used per default by \Webots), which means that there is no longer a fixed number of iterations, but rather a higher accuracy can be achieved at the expense of memory and computing time \cite{Drum2010}.

    Regarding the performance of our scenarios, we observed that neither of the simulation environments is able to scale successfully with multiple cores or an abundance of RAM.
    According to Tab.~\ref{tb:CPUusageRTF}, the CPU usage rarely ever went significantly over \SI{100}{\percent}, except for \Gazebo, thus mostly only utilizing one core.
    Consequently, the $RTF$ increased with the maximum frequency of the CPU itself.
    Moreover, we did not detect any use of the GPU whatsoever.
    However, we did not implement any visual sensors such as cameras, laser sensors or lidars, which could benefit from GPU acceleration \cite{Saglam2020}.
    Yet, there are simulation environments which promise performance increases through their use of GPU technology, see section \ref{ss:Future_Work}.
    Based on our results, the performance of the simulation environment is strongly dependent on the single core speed of the CPU.
    Thus, for a pure simulation workload with scenarios of similar scope and scale, we would recommend a high clocked CPU coupled with a medium tier GPU for rendering and a decent amount of RAM.
    However, this configuration does not take into account other use cases, such as the RL training process which can be conducted especially GPU bounded.
    
    The investigations carried out in this work refer to single simulation calculations. 
    We therefore want to indicate that an RL processes can be accelerated considerably with the help of parallelization. 
    Processors with a higher amount of logical cores will show advantages which, however, depend on the support of the simulation tool and the amount of parallel processes. 
    \Gazebo, \Pybullet and \Mujoco basically support parallel simulations. 
    In contrast, \Webots runs each simulation in the GUI and does not natively support parallelization. 
    Nevertheless, it is possible to open multiple \Webots instances and have them compute in parallel even though it requires a higher memory usage. 
    
    \subsection{Recommendations}
    \label{ss:recommendations}
    Based on our findings we would recommend each of the simulation environments in the following cases.
    \Gazebo would be preferred if one plans on developing not only for simulation but also for real systems.
    Due to the homogeneous ROS connection, one can use the same control interface for the simulation as well as the real physical system.
    While \Gazebo probably requires the steepest learning curve compared to other simulation environments, the close integration into the ROS framework does indeed offer its advantages.
    
    \Mujoco performed second-to-best for the lower number of degrees-of-freedom in the trajectory tracking of the UR10e but below-average handling the spheres.
    It can be an easy entry point for RL training as OpenAI Gym \cite{Brockman2016OpenAIGym} contains multiple working simulations for RL.
    However, other simulation environments offer similar functionalities based on a free and open-source concept. 
    
    \Pybullet is a well known open-source simulation tool with an extensive community.
    It enables even inexperienced users to find help and examples for their first simulation or RL projects.
    The simulation runs with a high amount of degrees-of-freedom showed better results than the low degrees-of-freedom robotic simulation case.
    Rather than speeding up the simulation runs by increasing the time steps which brings a strong, negative influence on the accuracy, we suggest to parallelize the simulations for the RL process.
    \Pybullet's client-based architecture enables the user to easily conduct several simulation runs parallel rather than increasing the time step.
    
    Finally, \Webots showed good simulation results even with high time steps for both few and many degrees-of-freedom scenarios.
    This enables the user to speed up their simulation without quality losses.
    While the GUI-based model setup offers easy and flexible customization of the models, the GUI-bound simulation complicates the parallelization of the simulation runs.
    \Webots enables a quick and easy model setup with good simulation results due to a customized physics engine which allows less parameter modifications compared to the other simulation environments.
    The predefined simulation parameter offer very good results out of the box.
    
    \subsection{Future Work}
    \label{ss:Future_Work}
    Besides the tested simulation environments, multiple others are currently developed.
    The most prominent being the Ignition \cite{Ignitionrobotics, Ferigo2020}, the successor of \Gazebo developed by OSRF.
    Compared to the monolithic architecture of \Gazebo, Ignition is build upon a modular system allowing the user to more easily change the physics engine, rendering engine or other components of Ignition.
    Developed by NVIDIA, Isaac Sim \cite{IsaacApp} as well as Isaac Gym \cite{NvidiaIsaacRelease}, currently both only available as early access versions, are meant to run on the CUDA cores of modern NVIDIA GPUs.
    Besides the promised performance improvements due to running the engine on the CUDA cores, Isaac Sim is also utilizing the new RTX cores to generate photorealistic images, especially useful for machine learning and RL.
    The support for a ROS API makes it a possible alternative for \Gazebo.
    Isaac Gym, on the other hand, is more targeted as an alternative for OpenAI Gym and \Mujoco.
    Lastly, RaiSim \cite{Hwangbo2018}, which is not yet released, also promises speed and accuracy improvements over the current state-of-the-art technology.
    Both Ignition and RaiSim were available in a stable and reliable release during the work on this paper, while Isaac Sim and Isaac Gym are still in development. 
    In the future those new simulation environments could be promising competitors to the tools discussed in this work.
%
\bibliographystyle{./bibliography/IEEEtran}
\bibliography{./bibliography/references}

\end{document}